\documentclass[a4paper,conference]{IEEEtran}

\IEEEoverridecommandlockouts

\usepackage{algorithm}
\usepackage{algpseudocode}
\usepackage{varwidth}
\usepackage{amsmath,epsfig}
\usepackage{amsfonts}
\usepackage{subfig}
\usepackage{enumitem}
\usepackage{balance}
\usepackage{color}

\graphicspath{{./images/}}
\DeclareGraphicsExtensions{.pdf,.jpeg, .png, jpg}

\newcommand{\hide}[1]{}

\newcommand{\minisection}[1]{\vspace{0.04in} \noindent {\bf #1}\ \ }

\begin{document}

\title{Robust pedestrian detection in thermal imagery using synthesized images}

\author{\IEEEauthorblockN{My Kieu\IEEEauthorrefmark{1}, Lorenzo Berlincioni\IEEEauthorrefmark{1}\thanks{\IEEEauthorrefmark{1}Both authors contributed equally to this research.}, Leonardo Galteri, Marco Bertini, Andrew D. Bagdanov, Alberto Del Bimbo}
\IEEEauthorblockA{%
  MICC - Università degli Studi di Firenze \\
  name.surname@unifi.it
}
}

\maketitle

\begin{abstract}
  In this paper we propose a method for improving pedestrian detection in the
  thermal domain using two stages: first, a generative data augmentation
  approach is used, then a domain adaptation method using generated data adapts
  an RGB pedestrian detector. Our model, based on the Least-Squares Generative
  Adversarial Network, is trained to synthesize realistic thermal versions of
  input RGB images which are then used to augment the limited amount of labeled
  thermal pedestrian images available for training. We apply our generative data
  augmentation strategy in order to adapt a pretrained YOLOv3 pedestrian
  detector to detection in the thermal-only domain. Experimental results
  demonstrate the effectiveness of our approach: using less than 50\% of
  available real thermal training data, and relying on synthesized data
  generated by our model in the domain adaptation phase, our detector achieves
  state-of-the-art results on the KAIST Multispectral Pedestrian Detection
  Benchmark; even if more real thermal data is available adding GAN generated
  images to the training data results in improved performance, thus showing that
  these images act as an effective form of data augmentation. To the best of our
  knowledge, our detector achieves the best single-modality detection results on
  KAIST with respect to the state-of-the-art.
\end{abstract}

%
\IEEEpeerreviewmaketitle

\begin{figure*}
  \includegraphics[width=\textwidth]{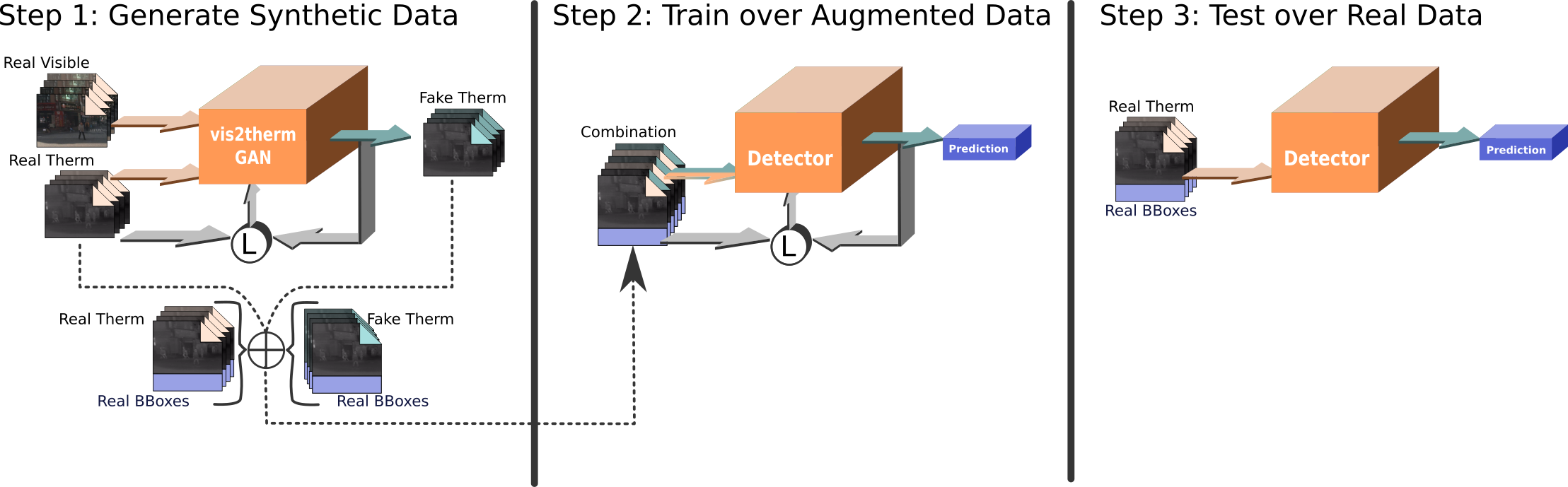}
  \caption{System overview: the vis2therm GAN generates fake thermal images from visible data; a mixture of real and fake thermal images along with related bounding boxes of objects are used to train an object detector, that is then tested on images from thermal cameras.}
  \label{fig:teaser}
\end{figure*}

\section{Introduction}

Pedestrian detection is a core problem in computer vision due to its central
role in a broad gamut of practical applications. Application areas such as
video surveillance and autonomous driving further require pedestrian detection
be robust across a range of illumination and environmental conditions, including
daytime, nighttime, rain, fog, etc. In such conditions, detectors based solely
on visible spectrum imagery can easily
fail~\cite{li2018multispectral,kieu2019domain}.

Detectors based on thermal imagery have garnered attention recently as a means
to mitigate the sensitivity of visible spectrum imagery to scene-incidental
imaging conditions~\cite{kieu2019domain, herrmann2018cnn, baek2017efficient}. A
growing number of works have also investigated multispectral detectors combining
visible and thermal images for robust pedestrian
detection~\cite{tian2015pedestrian,angelova2015real,wagner2016multispectral,konig2017fully,guan2019fusion,li2018multispectral,li2019illumination}.
Due to the cost of deploying multiple aligned sensors, multispectral models can
have limited applicability in real-world applications. Moreover, and especially
important given the recent focus on privacy by the public and national
legislative bodies, using visible spectrum sensors does not offer the same
privacy-preserving affordances as systems employing only thermal
sensors~\cite{kieu2019domain}.

Thermal-only detectors typically yield lower performance than multispectral
detectors since robust pedestrian detection using only thermal data is extremely
challenging. A key performance-limiting factor is the relative lack of annotated
thermal imagery available for training state-of-the-art models. Thermal
pedestrian datasets are few, and -- compared to visible-spectrum datasets -- have
orders of magnitude fewer annotated instances; for instance the Caltech
Pedestrian Dataset \cite{dollar2011pedestrian} has 350,000 annotations in the
visible domain, while KAIST Multispectral Pedestrian dataset
\cite{hwang2015multispectral} has $\sim 51,000$ annotations and FLIR ADAS
Dataset \cite{flir-adas} has $\sim 28,000$. Scaling thermal-only detection to
the levels of robustness and accuracy demanded by real-world applications is
thus extremely difficult due to this poverty of annotated data.

In this paper we propose to use a generative algorithm to perform data
augmentation that can enrich thermal pedestrian datasets for training deep
detector architectures. Our approach is based on a Least-Squares Generative
Adversarial Network (LSGAN)~\cite{DBLP:journals/corr/MaoLXLW16} trained to
synthesize thermal pedestrian images from RGB inputs. We investigate the best
approaches to exploit these generated images during training, i.e.~studying how
to mix real thermal images with synthesized ones in order to effectively augment
the training set. Experimental results indicate that our trained LSGAN is able
to learn to translate RGB pedestrian images to useful thermal versions so that
even using $\sim 50\%$ synthetic images results in state-of-the-art pedestrian
detection at nighttime and overall day/nighttime. This suggests that the
approach can be extended to other domains in which thermal training data is
scarce but is possible to effectively exploit the abundance of RGB imagery to
adapt it to the thermal domain.

The contributions of this work are:
\begin{itemize}
\item we propose a novel generative model based on the Least-Squares Generative
  Adversarial Network (LSGAN)~\cite{DBLP:journals/corr/MaoLXLW16} that is able
  to synthesize thermal imagery from RGB;
\item we propose a mixed real/synthetic training domain adaptation procedure
  that mixes real thermal imagery with thermal images synthesized from unlabeled
  RGB pedestrian images using our LSGAN and uses this augmented training set to
  adapt the YOLOv3 \cite{redmon2017yolo9000} detector;
\item we conduct extensive ablation study to probe the effectiveness of our
  approach and a variety of mixing proportions of real and synthesized imagery;
  and
\item we conduct an extensive set of experiments comparing our approach to the
  state-of-the-art, and to the best of our knowledge our thermal-only detector
  outperforms all state-of-the-art single-modality detection approaches on the
  KAIST Multispectral Pedestrian Detection
  Benchmark~\cite{hwang2015multispectral} by a large margin.
\end{itemize}

The rest of the paper is organized as follows. In the next section we review the
scientific literature related to our proposed approach. In
section~\ref{sec:proposed_system} we describe our generative model used to
synthesize thermal images and our training procedure used to adapt a YOLOv3
pedestrian detector to the thermal domain. We report in
section~\ref{sec:experiments} on an extensive set of experiments performed to
evaluate the effectiveness of thermal pedestrian detection using our approach,
and in Section~\ref{sec:conclusions} we conclude with a discussion of our
contribution.

\section{Related work}
The problem of pedestrian detection in thermal imagery has attracted much
attention from the research community over the years due to the advantages of
thermal cameras in many real-world and critical applications.

\subsection{Pedestrian detection in thermal imagery}
Thanks to the reduction of costs and availability of multispectral cameras over the
past few years, there are numerous recent works exploiting thermal images in
combination with visible images for robust pedestrian detections~\cite{wagner2016multispectral,liu2016multispectral,konig2017fully,xu2017learning,li2018multispectral,li2019illumination,fritz2019generalization,zhang2019cross,cao2019box,vandersteegen2018real,lee2018pedestrian,zheng2019gfd}.
In contrast, many recent works have investigated pedestrian detection using
thermal (IR) imagery only. For example, authors in~\cite{john2015pedestrian}
used Adaptive fuzzy C-means for IR image segmentation and a CNN for pedestrian
detection. In~\cite{baek2017efficient} the authors proposed a combination of
Thermal Position Intensity Histogram of Oriented Gradients (TPIHOG) and the
additive kernel SVM (AKSVM) for nighttime-only detection in thermal imagery.
Thermal images augmented with saliency maps, used as attention mechanism, have
been used in~\cite{ghose2019pedestrian}.

The idea of performing several video preprocessing steps to make thermal images
look more similar to grayscale images converted from RGB was investigated
in~\cite{herrmann2018cnn}, who then applied a pretrained and fine-tuned SSD
detector. Recently, authors in~\cite{9064036} designed dual-pass fusion block
(DFB) and channel-wise enhance module (CEM) to improve the one-stage detector
RefineDet, and proposed their ThermalDet detector for pedestrian detection in
thermal imagery. Another recent single-modality work was the Bottom-up Domain
Adaptation approach proposed in~\cite{kieu2019domain} for pedestrian detection
in thermal imagery. We also focus on the thermal-only detection problem.
However, our approach is distinct in that we concentrate on domain adaptation
via data augmentation during training using synthetic thermal data which is
generated by a generative model trained on unlabeled data.

\subsection{Spectrum transfer between visible and thermal}

The generation of RGB images from the thermal images has been approached as a
grayscale colorization task in several previous works such as
\cite{DBLP:journals/corr/LimmerL16} where deep multiscale CNNs are used along
with classical computer vision post processing techniques over near infrared
images. In~\cite{colTherm2018} a CNN is used with a more sophisticated objective
function in order to tackle misalignment issues between the two visible and
thermal modalities. In~\cite{dong2018infrared} instead an encoder-decoder
architecture is applied for performing colorization.

Most recent works, however, rely heavily on generative models to
perform image-to-image translation between visible and thermal.
As defined in~\cite{isola2017image}, the \textit{image-to-image translation}
problem is the task of translating one visual representation of a scene
into another. Many \textit{domain to domain} translation problems
\cite{Choi2017StarGANUG}, from image denoising \cite{elad2006image} to image
super-resolution \cite{nasrollahi2014super}, can be cast as image-to-image
translation tasks.

Generative Adversarial Networks (GANs), introduced
in~\cite{goodfellow2014generative}, are one the most significant recent
improvements in the field of generative models and have been extensively used for
image-to-image translation. The key feature of these models is the
competitive min/max game between two networks. GANs have been successfully
applied in many computer vision tasks such as super resolution
\cite{galteri2017deep,ledig2017photo,wang2018esrgan}, style
transfer\cite{CycleGAN2017}, image inpainting
\cite{DBLP:journals/corr/YehCLHD16} and domain
adaptation\cite{Hoffman2018CyCADACA}.

Both \cite{suarez2018,Mehri_2019_CVPR_Workshops} use GANs architectures to
perform infrared and grayscale colorization. In \cite{suarez2018} a DCGAN with
one seperate generator per channel is used, while in
\cite{Mehri_2019_CVPR_Workshops} an improved \cite{CycleGAN2017} GAN is
proposed. In \cite{wang2018} the authors focused on learning an
identity-preserving translation between thermal and visible images of faces. The
authors in~\cite{polarimetric2019} leverage multiple streams of polarimetric
images to synthesize photo-realistic visible images of faces preserving
discriminative features. In \cite{in2I} a multi-image to image generative
framework is presented, and one of the proposed settings is
infrared and grayscale colorization. Also in~\cite{devaguptapu2019borrow} the
authors used a Cycle-GAN\cite{CycleGAN2017} for image-to-image translation of
thermal to pseudo-RGB data. The use of these frameworks to perform data
augmentation in order to improve the performance of a seperate classifier has
been studied in multiple previous works such as
\cite{10.1007/978-3-030-01424-7_58} in which they focus on improving one-shot
learning, in \cite{Bowles2018GANAA} where segmentation of medical images is
enhanced by GAN augmented data.


In this work we focus on the opposite task: mapping RGB images to the infrared
spectrum. The closest related works are \cite{DBLP:journals/corr/abs-1812-08333,
  zhang-2019, Guo2019DomainAdaptivePD, 10.1007/978-3-030-11024-6_46}, as they
all employ generative models to translate images from the visible to the thermal
spectrum. A modified Cycle-GAN~\cite{CycleGAN2017} is used in
\cite{DBLP:journals/corr/abs-1812-08333}, where the performance of drone
detection in the thermal spectrum is improved using augmented data coming from a
visible to thermal GAN framework, and also in \cite{Guo2019DomainAdaptivePD},
where a pedestrian detector is trained on augmented thermal data. Also
in~\cite{DBLP:journals/corr/abs-1812-08333} a modified version is proposed which
changing the loss with a perceptual texture loss term. In~\cite{zhang-2019},
both pix2pix~\cite{isola2017image} and Cycle-GAN are used to generate thermal
images to train an object tracker in the thermal domain; experiments show that
images generated with pix2pix are of higher quality, since this approach
operates on paired thermal/RGB data.

The authors of~\cite{10.1007/978-3-030-11024-6_46} present a framework for
cross-modality color to thermal person re-identification. The generative model
in this work is tasked with the generation of multiple thermal versions of the
visible input image, which is then used to match with real thermal gallery set.
Here the proposed architecture is a variation of~\cite{bycyclegan}, a multimodal
image-to-image translation framework composed of multiple networks: cVAE-GAN
from~\cite{pmlr-v48-larsen16} and cLR-GAN from~\cite{infogan} which are jointly
optimized in a hybrid model in order to cover complementary tasks. One of the
major contribution of~\cite{bycyclegan} is the ability to model the distribution
of different correct outputs corresponding to the same input.

In our approach we instead rely on a different architecture that combines
elements from \cite{DBLP:journals/corr/MaoLXLW16} and \cite{wang2018esrgan}, as
further detailed in Section~\ref{subsec:visible_to_thermal_GAN}. The proposed
architecture in \cite{wang2018esrgan}, ESRGAN, focuses on the
\textit{super-resolution} problem and improved over the previous
state-of-the-art \cite{DBLP:journals/corr/LedigTHCATTWS16} by introducing the
Residual-in-Residual Dense Block, removing the Batch-Normalization layers, and
changing the perceptual loss term.


\section{Generative data augmentation for thermal domain adaptation}
\label{sec:proposed_system}

In this section we describe the two main components of our proposed approach.
Our thermal pedestrian detector based on YOLOv3~\cite{redmon2018yolov3} is
described in the next section, and our generative model which produces fake
thermal images from available RGB images is described in
section~\ref{subsec:visible_to_thermal_GAN}. An extensive series of experimental
results are reported on in section~\ref{subsec:detection_ablation_studies}.

\subsection{Object detection in thermal images}
\label{subsec:detection_thermal_video}

\begin{figure}
  \centering
  \includegraphics[width=\linewidth]{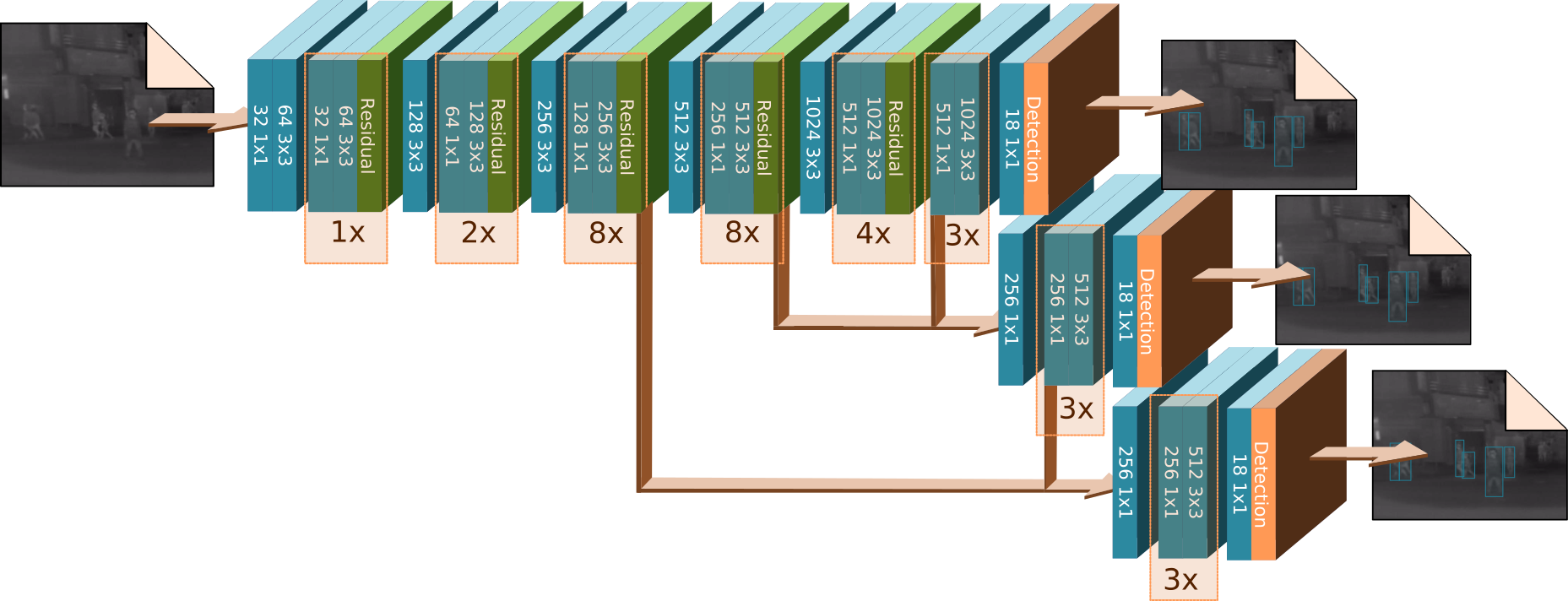}
  \caption{The YOLOv3 architecture. $k\times$ indicates the repetition of blocks $k$ times.}
  \label{fig:yolov3_architecture}
\end{figure}

We use YOLOv3 as our base pedestrian detector~\cite{redmon2018yolov3}. Following
the Domain Adaptation approach described in~\cite{kieu2019domain}, we first
adapt YOLOv3 in the visible domain by directly fine-tuning it on the visible
spectrum images from the KAIST dataset~\cite{hwang2015multispectral}. Then, we
use this detector as a starting point for training a thermal detector using a
range of mixtures of real and GAN-generated thermal images.
Figure~\ref{fig:yolov3_architecture} illustrates the original YOLOv3
architecture with thermal image as input and the output of the model at three
detection scales.

We consider the following training regimes for thermal detectors:
\begin{itemize}
\item \textbf{Real-Thermal detector}: We directly fine-tune the detector on all
  available \emph{real thermal images}.
\item \textbf{Synthesized-Thermal detector}: We directly fine-tune the detector
  on all the \emph{GAN-generated thermal images (synthesized images)}.
\item \textbf{Combined-Thermal detector}: We combine all available real images and all
  the synthesized images into a combined training set and then we fine-tune the
  detector on it. Note that the number of images in this combined set is
  double that used for the Real-Thermal and Synthesized-Thermal detectors.
\item \textbf{Mixed-Thermal detectors}: We mix real images and synthesized
  images with a proportion varying from 10\% to 90\%; in total we have 9 mixed
  sets of images. For example, the mixed set 1 has 10\% real images and 90\%
  synthesized images. Note that the number of images used to train these
  detectors is the same as those used for Real-Thermal and Synthesized-Thermal
  detectors. 
\end{itemize}
For all experiments we evaluate performance on the KAIST test set of real
thermal images.

\subsection{Visible to thermal GAN}

\label{subsec:visible_to_thermal_GAN}
\begin{figure*}
  \centering
  \begin{tabular}{cc}
      \includegraphics[height=0.75in]{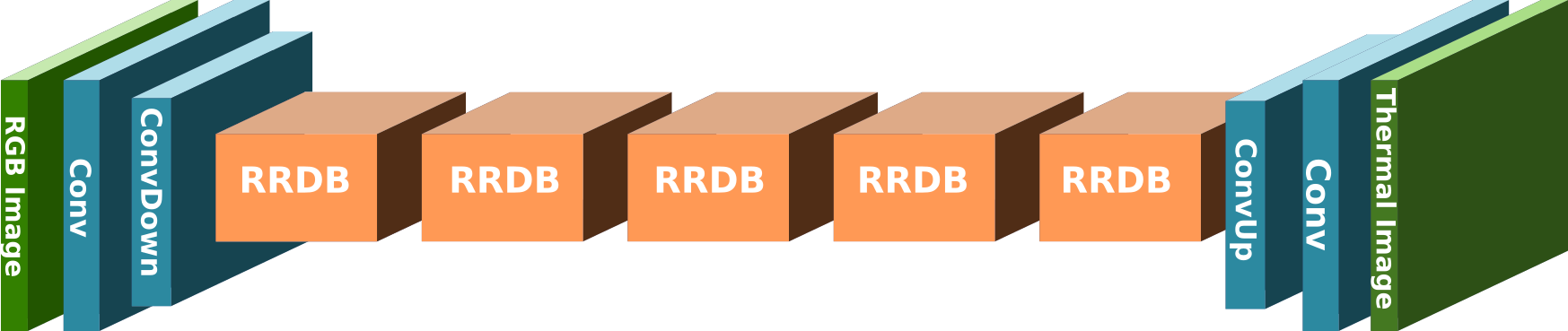}
    & \includegraphics[height=1.2in]{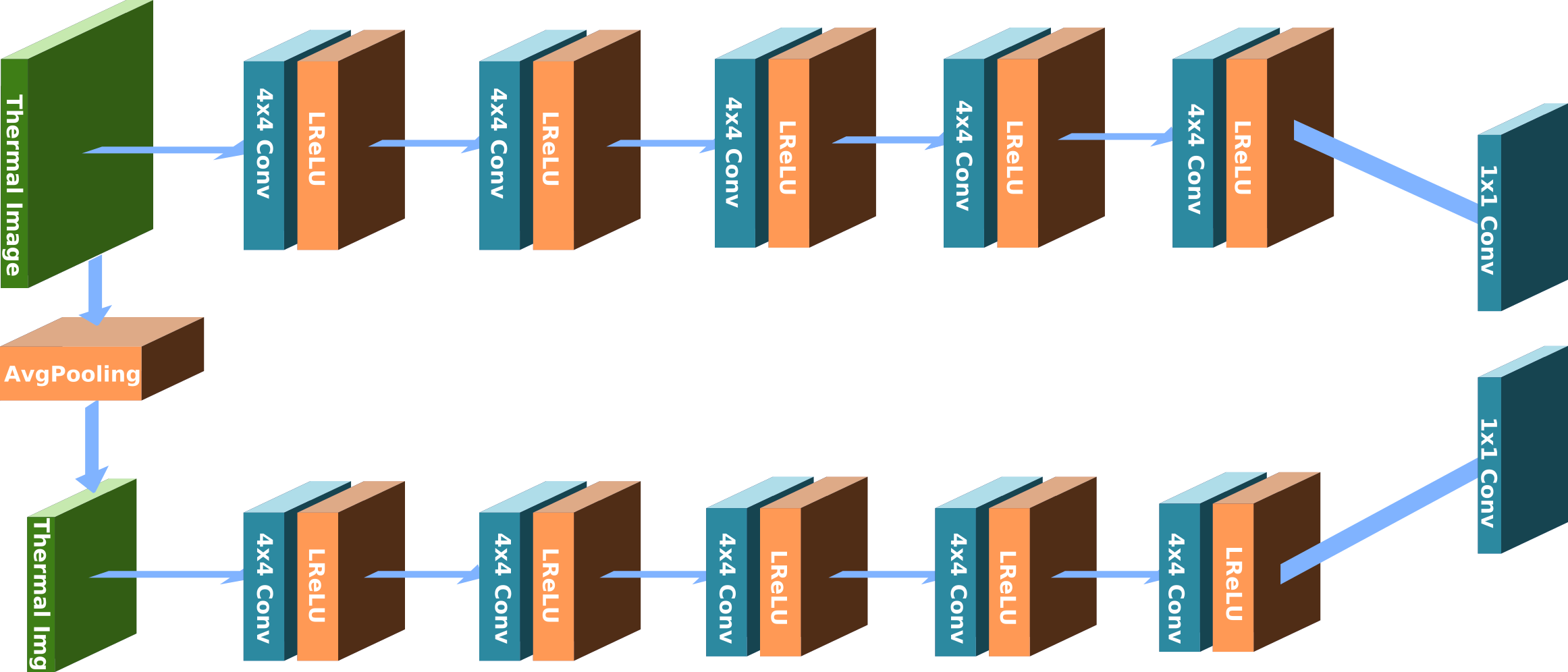} \\
    (a) Generator architecture & (b) Discriminator architecture
  \end{tabular}
  \caption{Our LSGAN architecture. The prosposed generator (a), composed of multiple Residual in Residual Dense Blocks, and the discriminator structure (b), a multiscale CNN.}
  \label{fig:gen_arch}
\end{figure*}

\begin{figure}
  \centering
  \includegraphics[width=\linewidth]{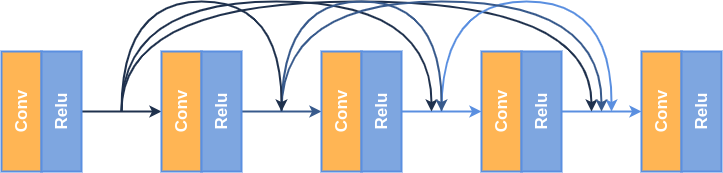}
  \caption{Dense Blocks. Arcs represent the concatenation between the output of a layer and one of its subsequent layers.}
  \label{fig:dense_block}
\end{figure}

\begin{figure}
  \centering
  \includegraphics[width=\linewidth]{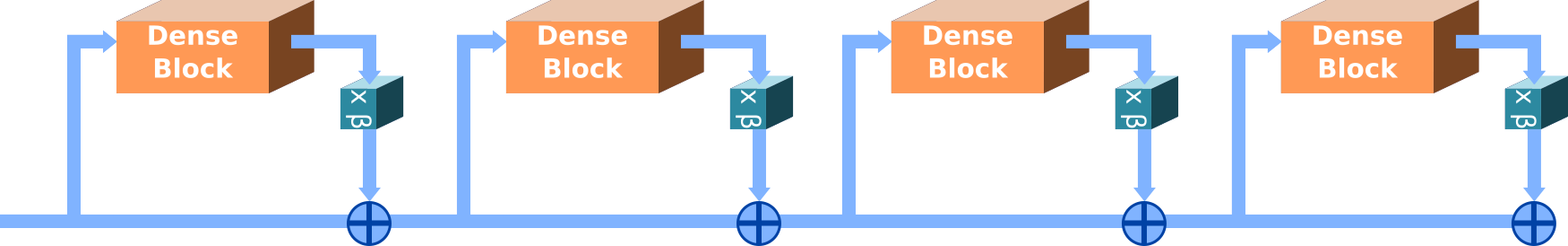}
  \caption{Residual in Residual Dense Block. The output of Dense Blocks are
    scaled by $\beta$ and summed back to their input.}
   \label{fig:RRDB}
\end{figure}

Our model is an LSGAN trained with both Adversarial and Perceptual losses. The
Least Squares GAN (LSGAN)~\cite{DBLP:journals/corr/MaoLXLW16, Mao_2017_ICCV}
improves on the standard GAN model by changing the loss function from a
cross-entropy to a squared distance. It is comparatively more stable and easier
to train. The Generator $G$ architecture is built using the Residual in Residual
Dense Block (RRDB) as the fundamental unit (see Figure~\ref{fig:RRDB}).
As in \cite{Lim2017EnhancedDR}, we remove the batch normalization layer from the traditional
\textit{Conv-BN-LReLU} triplet. After the initial down-sampling convolutions five
RDDB blocks are stacked in sequence as shown in Figure~\ref{fig:gen_arch}(a). Each
RDDB block is composed of 4 Dense Blocks. Each Dense Block has a
growth rate of $k = 32$ and contains five consecutive pairs of convolutional
layers followed by a leaky rectified linear unit (LReLU) whose outputs are
concatenated as shown in Figure~\ref{fig:dense_block}.

\minisection{Dense Blocks.} DenseNets, introduced in
\cite{DBLP:journals/corr/HuangLW16a}, improve the information flow between layers
by adding direct connections between a layer and all subsequent layers. By
using this connectivity pattern the $l^{th}$ layer receives the feature maps
coming from all the preceding $l-1$ layers as shown in
Fig.~\ref{fig:dense_block}. This dense connection strategy is realized by
feeding as input the concatenation of every preceding layer output. DenseNets
provide advantages both from a memory consumption and a vanishing gradient
standpoint.

\minisection{Residual in Residual Block.} The composition of Residual
Networks~\cite{DBLP:journals/corr/HeZRS15} and DenseNets is the Residual in
Residual Dense block (RRDB), as introduced in \cite{wang2018esrgan}. A single
RRDB is composed of multiple Dense blocks connected in a residual fashion, and
is shown in Fig.~\ref{fig:RRDB}. Finally, the output of the RRDB chain is
followed by multiple \textit{upscale-Conv-ReLU} blocks to scale the image back
to input size.

Inspired by \cite{wang2018high,DBLP:journals/corr/DurugkarGM16,Karnewar_2020_CVPR} successful application of multi scale architectures we use a multi-scale discriminator $D$, shown in Figure~\ref{fig:gen_arch}(b), that makes no use of dense connectivity patterns. It is
composed of five convolutional layers, each of them using a $4 \times 4$
convolutional kernel with stride 2 and followed by LReLU activation function.
The number of feature maps is doubled as depth increases starting from 64. For
each of the multiple scales, a single $1 \times 1$ convolutional filter is used
as final output layer. Finally, the different outputs of every scale is evaluated
independently.

\minisection{Training.} We trained the model as a Least Squares
Generative Adversarial Network (LSGAN) with a perceptual loss. The discriminator $D$ is
trained as a standard LSGAN Discriminator:
\begin{eqnarray*}
  L_{D_{LSGAN}} &=& \frac{1}{2} \mathbb{E}_{x \sim p_{data}(x)}[(D(x)-real_{label})^2] \\
                    &+& \frac{1}{2} \mathbb{E}_{z\sim p(z)}[(D(G(z))-fake_{label})^2].
\end{eqnarray*}

The generator loss is composed of three terms:
\begin{eqnarray}
\label{eq:lgadv}
  L_{G_{Adv}}&=&\frac{1}{2} \mathbb{E}_{z \sim p(z)}[(D(G(z))-{fake_{label}})^2] \\
  \label{eq:lgmae}
  L_{G_{MAE}} &=& |real_{img}-fake_{img}| \\
  \label{eq:lgper}
  L_{G_{Perceptual}} &=& (\phi^{k}(real_{img})-\phi^{k}(fake_{img}))^2,
\end{eqnarray}
which are summed together:
\begin{equation}
  \label{eq:loss}
  L_{G_{LSGAN}}=L_{G_{Adv}}+L_{G_{MAE}}+L_{G_{Perceptual}}
  \end{equation}  

\minisection{Perceptual loss.}
\label{perp_loss}
Perceptual loss functions~\cite{johnson2016perceptual} aim to provide a better
measure for similarity compared to metrics such as the PSNR (Peak Signal to
Noise Ratio) and SSIM (Structural Similarity Index). They have been shown
useful for super-resolution and style-transfer tasks. Our perceptual loss 
architecture consists of two networks:
 \begin{itemize}
     \item Transformation Network $T$
     \item Loss Network $\phi$
\end{itemize}
The Loss Network $\phi$ is pretrained, usually as a classifier. When training
the transformation network $T$, the loss network $\phi$ is used as a feature
extractor by taking the output of some of its layers. The distance between the
target and the generated image in this feature space is used as a loss function
for the Transformation Network $T$. The main motivation behind perceptual loss
functions lies on the intuition that computing distances in the high dimensional
manifold extracted from a well-trained classifier should result in a better
estimate compared to any pixel-space distance measure.

As shown in \cite{NIPS2016_6158} pixel space metrics can lead to minima that
corresponds to blurry results. In this work, since our goal is to detect
pedestrians, we use the YOLO detector to drive the generation of the images. The
term (\ref{eq:lgper}) is a \textit{perceptual loss} defined as the squared
distance between the outputs $\phi^{k}$ of the $k^{th}$ layer of a pretrained
YOLOv3 network for a real and a generated input. We trained the $\phi$ network
on KAIST for a detection task in a thermal setting. We choose the last
convolutional layer of YOLOv3 as representation of the input image in the high
dimensional space learned by the classifier. Note that the loss network $\phi$
at this stage acts as a feature extractor and its weights are frozen.

\section{Experimental results}
\label{sec:experiments}
In this section we report on a range of experiments conducted to evaluate the
effectiveness of our approach to thermal domain adaptation for pedestrian
detection. We first describe the dataset and evaluation metrics used, then in
Section~\ref{subsec:GAN_ablation_results} give a qualitative evaluation of the performance
of our GAN in generating thermal imagery from RGB input. In
Section~\ref{subsec:detection_ablation_studies} we perform an ablative analysis
of the use of synthetically generated thermal imagery for data augmentation, and
in Section~\ref{subsec:object_detection_results} give a comparison with the
state-of-the-art.

\subsection{Dataset and experimental protocol}

\minisection{Dataset.}
\label{subsec:dataset}
All of our experiments were conducted on the KAIST Multispectral Pedestrian
Benchmark dataset~\cite{hwang2015multispectral}. KAIST is a large-scale dataset
with well-aligned visible/thermal pairs~\cite{devaguptapu2019borrow}, and it
contains videos captured both during the day and at night. KAIST dataset
consists of 95,328 image pairs split into 50,172 for training and 45,156 for
testing. We follow the standard sampling procedure
in~\cite{hwang2015multispectral,li2018multispectral,liu2016multispectral}, we
sample every two frames from training videos and exclude heavily occluded and
small person instances ($<50$ pixels). The final training set contains 7,601
images. The test set contains 2,252 image pairs sampled every 20 frames. For
training and testing, we use the improved training annotations
from~\cite{li2018multispectral} and test annotations
from~\cite{liu2016multispectral}.

\minisection{Performance metrics.}
As is common practice to compare with the state-of-the-art, we used standard
evaluation metrics for object detection, namely miss rate as a function of False
Positives Per Image (FPPI), and log-average miss rate for thresholds in the
range of $[10^{-2} , 10^0]$ with an Intersection over Union (IoU) threshold of
0.5 under the \emph{reasonable}
setting~\cite{dollar2011pedestrian,hwang2015multispectral,li2018multispectral,liu2016multispectral,kieu2019domain}.
The \emph{reasonable} setting is composed of \emph{day-time}, \emph{night-time},
and \emph{all (both day and night time)} sets of images.
Figure~\ref{fig:kaist_example} shows some example images with our detection
results on KAIST dataset.

\minisection{Fine-tuning.}
\label{minisection:finetuning}
All of our detectors were implemented using PyTorch. During fine-tuning to adapt
to the thermal domain, at each epoch we set aside 10\% of the training images
for validation for that epoch. We trained every detector using Stochastic
Gradient Descent with the same procedure and hyperparameters: image size
$640 \times 512$, batch size of 4, We set an initial learning rate of 0.001 if
the training set contains 50\% or more real images, otherwise we use a learning
rate of 0.0001. During fine-tuning, we reduce the learning rate by a factor of
10 every 3 epochs, and training is halted after 10 epochs.

\subsection{GAN results}
\label{subsec:GAN_ablation_results}

\begin{figure*}
    \centering
  \includegraphics[width=0.3\textwidth,trim={0 50px 200px 150px},clip]{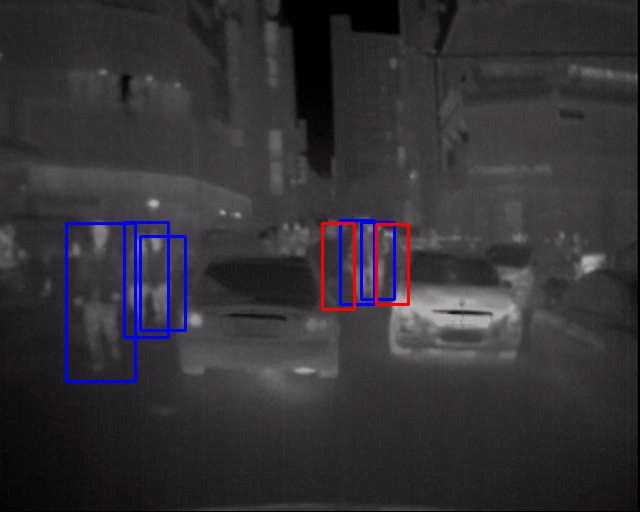}
  \includegraphics[width=0.3\textwidth,trim={0 50px 200px 150px},clip]{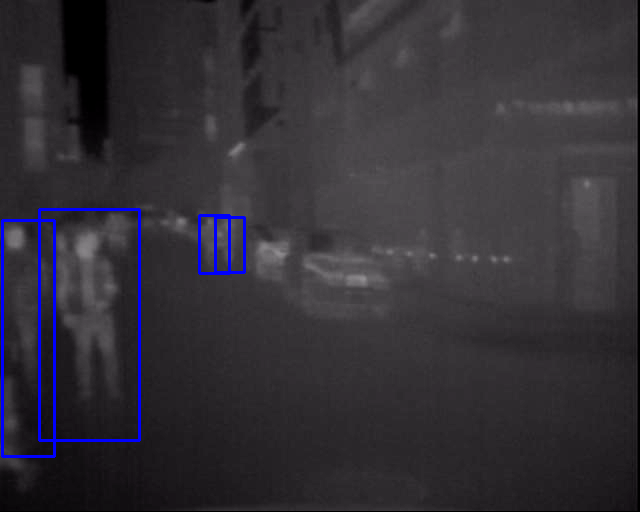}
  \includegraphics[width=0.3\textwidth,trim={150px 100px 50px 100px},clip]{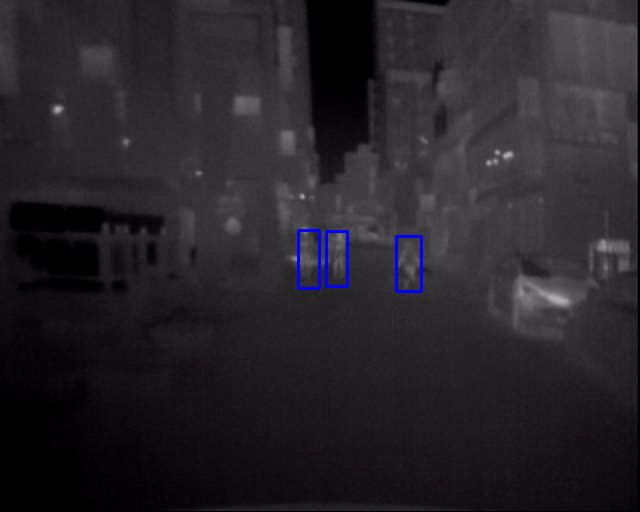}
  \\
  \includegraphics[width=0.3\textwidth,trim={0 50px 200px 150px},clip]{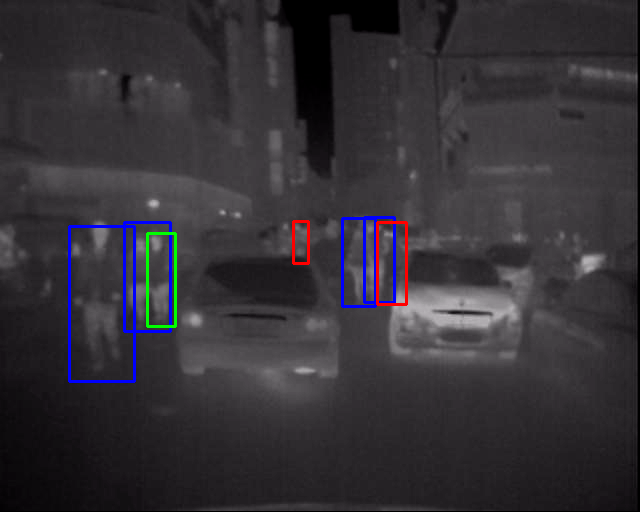}
  \includegraphics[width=0.3\textwidth,trim={0 50px 200px 150px},clip]{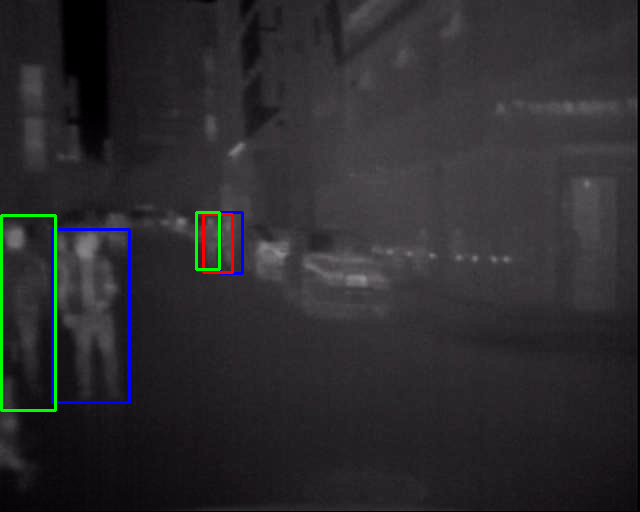}
  \includegraphics[width=0.3\textwidth,trim={150px 100px 50px 100px},clip]{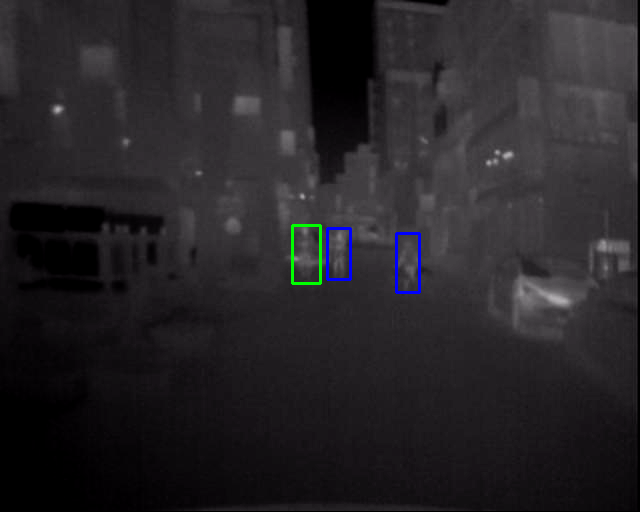}
  \caption{Example detections using the detector trained with 80\% real images
    and 20\% synthesized images. The first row shows detection results with the
    perceptual loss, while the second row is \emph{without} perceptual loss.
    \textcolor{blue}{Blue boxes} are \textcolor{blue}{true positive detections},
    \textcolor{green}{green boxes} are \textcolor{green}{false negatives}, and
    \textcolor{red}{red boxes} indicate \textcolor{red}{false positives}}
  \label{img:detection_mix20_perc_and_without}
\end{figure*}

The GAN framework for the \textit{visible to thermal} transformation was trained
on pairs of RGB-LWIR frames from the original training split of the KAIST
dataset. In Figure~\ref{img:detection_mix20_perc_and_without} we show some
examples detections using the detector trained with 20\% sythesized images and
80\% real images on two kinds of images. The first row shows detection results
on generated images without Perceptual Loss $L_{G_{Perceptual}}$, and the second
row gives detection results on generated images by our model trained with
$L_{G_{Perceptual}}$. The use of the $L_{G_{Perceptual}}$ seems to result in
more true positive (blue boxes) detection results, as well fewer false negative
(green boxes).

\setlength{\tabcolsep}{4pt}
\begin{table}
  \centering
  \caption{Ablation study on varying quantities of GAN-generated images. Results are on KAIST in terms of log-average miss rate
    (lower is better). Best results highlighted in \textbf{\underline{underlined bold}},
    second best in \textbf{bold}.}

  \label{tab:ablation_studies}
  \begin{tabular}{|r|c|c|c|c|c|}
    \hline
 & \multicolumn{2}{|c|}{\textbf{Mixture}} & \multicolumn{3}{c|}{\textbf{Miss Rate (\%)} }\\
    \hline \hline
 & Real (\%) & Synthetic (\%)& all & day   & night \\
    \hline
\textbf{Synthesized} & 0 & 100	& 45.88 & 54.37 & 26.04	 \\
\hline
& 10 & 90	& 44.90 & 54.24 & 22.79	 \\
& 20 & 80	& 41.21 & 51.04 & 18.92 \\
& 30 & 70	& 35.32 & 44.44 & 16.35 \\
& 40 & 60	& 34.78 & 43.45 & 14.53	 \\
\textbf{Mixed} & 50 & 50	& 33.90 & 41.97 & 14.64 \\
& 60 & 40	& 31.50 & 39.83 & 12.33 \\
& 70 & 30	& 32.29 & 41.68 & 12.42 \\
& 80 & 20	& \bfseries 25.88 & \bfseries 33.01 & \bfseries \underline{11.12} \\
& 90 & 10	& \bfseries \underline{25.62} & \bfseries \underline{31.86} & 12.92  \\
\hline
\textbf{Real} & 100 & 0	& 28.46 & 36.32 &  \bfseries 11.97 \\ \hline
\textbf{Combined} & all & all & 34.29 & 41.93 & 16.80  \\
    \hline
  \end{tabular}
\end{table}

\setlength{\tabcolsep}{1.4pt}

\subsection{Ablation study}
\label{subsec:detection_ablation_studies}
In this section, we report on a series of experiments we conducted to explore
the many options available when using GAN generated images (synthesized images)
and thermal images (real images) for training the detectors described in
Section~\ref{subsec:detection_thermal_video}. 
Initial experiments with simple augmentation strategies resulted in worse results than the conventional fine-tuning model. Thus, we use the conventional fine-tuning result as a baseline for comparison with various mixing strategies of GAN-generated thermal images.
In table~\ref{tab:ablation_studies} we present results of an ablation study
considering all these possibilities. From these results we first note that
mixing in a \emph{small} proportion of synthesized images (\textbf{Mixed})
rather than training on a all available real and synthesized images
(\textbf{Combined}) is generally useful. In fact, the best mixture proportion is
90\% real images with 10\% percent synthesized images with 25.62\% miss rate the ``all" setting, and the second best is the \textbf{Mixed} of 80\% and 20\% with 11.12\% miss rate in nighttime -- an improvement of 5.68\% over the \textbf{Combined} using all
available data. Note that even with fewer
than 50\% real images our detector achieves results are comparable with
state-of-the-art methods. Moreover, observe that mixing more than 50\% real
images results in improvement over the detector that combining all available
real and synthesized images. 
The result reveals that the small portion of GAN synthesized images is useful for augmentation approach, but it must be consider based on the testing data such as the real test set was conducted on the test phase, thus the \textbf{Mixed} and \textbf{Real} results are better a little than the \textbf{Combined} result.

\subsection{Comparison with the state-of-the-art}
\label{subsec:object_detection_results}

\setlength{\tabcolsep}{4pt}
\begin{table}
  \centering
  \caption{Comparison with state-of-the-art single-modality approaches
    on KAIST Thermal in term of log-average miss rate (lower is better). Best
    results highlighted in \textbf{\underline{underlined bold}},
    second best in \textbf{bold}.}
  \label{tab:compare_single_modality}
  \begin{tabular}{|l|c|c|c|}
    \hline
    \textbf{Detectors} & \textbf{MR all}   & \textbf{MR day}
    & \textbf{MR night} \\
    \hline \hline
    KAIST baseline \hfill \cite{hwang2015multispectral} & 64.76 & 64.17 & 63.99 \\
    FasterRCNN \hfill \cite{liu2016multispectral}      & 47.59 & 50.13 & 40.93 \\
    TPIHOG \hfill \cite{baek2017efficient} & - & - & 57.38 \\
    SSD300 \hfill \cite{herrmann2018cnn} &  69.81 & - & -  \\
    Saliency + KAIST \hfill \cite{ghose2019pedestrian} & - & 39.40 & 40.50  \\
    $R^3$-Net Saliency + KAIST \hfill \cite{ghose2019pedestrian} & - & \bfseries \underline{30.40} & 21.00  \\
    VGG16-two-stage \hfill \cite{Guo2019DomainAdaptivePD} & 46.30 & 53.37 & 31.63  \\
    ResNet101-two-stage \hfill \cite{Guo2019DomainAdaptivePD} & 42.65 & 49.59 & 26.70 \\
    Bottom-up \hfill \cite{kieu2019domain} & 35.20 & 40.00 & 20.50 \\
    \hline \hline
    \textbf{Ours} Mixed 40\_60	& 34.78 & 43.45 & 14.53	 \\
    \textbf{Ours} Mixed 80\_20 & \bfseries 25.88 & 33.01 & \bfseries \underline{11.12}\\
    \textbf{Ours} Mixed 90\_10 & \bfseries \underline{25.62} & \bfseries 31.86 & \bfseries 12.92\\
    \hline
  \end{tabular}
\end{table}
\setlength{\tabcolsep}{1.4pt}

\begin{figure*}[t]
  \centering
  \includegraphics[width=\textwidth]{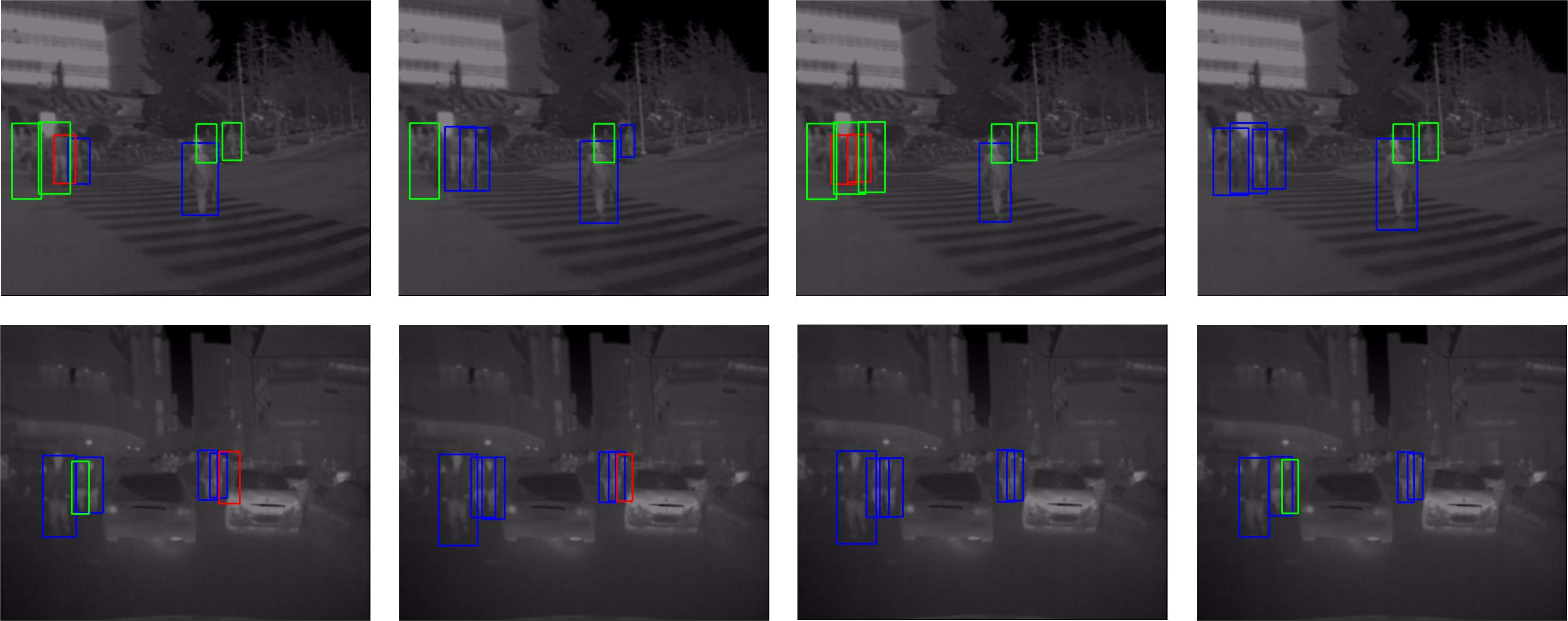}
  \caption{
  Examples of KAIST thermal images with detections. The first row is daytime images and the second is nighttime. The first and the second column are detection result on synthetic-only and real-only training, respectively. The third and the last column are combining all and mixed 90\% proportion, respectively. \textcolor{blue}{Blue boxes} are
    \textcolor{blue}{true positive detections},
    \textcolor{green}{green boxes} are \textcolor{green}{false
      negatives}, and \textcolor{red}{red boxes} indicate
    \textcolor{red}{false positives}. See section~\ref{subsec:object_detection_results}
    for detailed analysis.}
  \label{fig:kaist_example}
\end{figure*}

Table~\ref{tab:compare_single_modality} compares our results with the state-of-the-art single modality approaches which are mostly trained and tested only on thermal images of KAIST dataset (except the KAIST baseline~\cite{hwang2015multispectral} that is a  multispectral method), some other models also used visible images for transfer learning such as ~\cite{kieu2019domain}. We leveraged unlabeled RGB images of train set for generating synthetic thermal images, then we used this thermal data as augmentation for training; of course, testing was conducted on real thermal images of the test set. Results are compared in terms of log average miss rate (lower score is better). We can see that our approaches obtained the best results with 25.62\% of missrate at ``all" and 11.12\% of missrate at ``nighttime" -- an improvement of 9.38\% over the second state-of-the-art results. Moreover, our results outperform all existing the state-of-the-art methods by a large margin in both ``night-time" and ``all".
The results of $R^3$-Net Saliency~\cite{ghose2019pedestrian} are a little better than ours in day time due to the advantages of their proposed pixel-level ``saliency'' annotation set with manually annotated 1,702 images from training and 369 from testing set, and their extraction of deep saliency maps by $R^3$-Net for augmenting thermal images of both training and testing.

Several different backbones have been used by the methods reported in the table, from VGG16 to Faster RCNN. Our backbone is the conventional YOLOv3 detector, and as fine tuning procedure  we followed our previous approach of~\cite{kieu2019domain}. The improvements that allowed to surpass the second-best state-of-the-art detector on KAIST (bottom-up~\cite{kieu2019domain}) are: 1) the new data annotation as described in section~\ref{subsec:dataset}; 2) the domain adaptation method of~\cite{kieu2019domain} and the experimentation with hyperparameter setting reported in section~\ref{minisection:finetuning}. Moreover, with the proposed generated synthesized thermal images with LSGAN and the mixed training procedure, we achieve state-of-the-art performance for both all (day and night)
and nighttime. 

It is expected that detection in thermal images at nighttime will
always be better than daytime results because of the low contrast between
pedestrians and background during the day, as noted
in~\cite{ghose2019pedestrian}.



In Figure~\ref{fig:kaist_example} we show some example detections from four detectors (synthetics, real, combination and mixed90). From these
examples we see that the mixed of 90\% real images with 10\% synthesized images yields more true positive and
fewer false positive detections with respect to others.
Not surprisingly, \textbf{synthesized detector} (the first column)  produces a higher number of false positives and missed detections than \textbf{real detector} (the second column). The difference is even more pronounced at nighttime (second row of figure~\ref{fig:kaist_example}). The mixed scale 90\% real with 10\% synthesized images for training (the last columns) makes more true positive and less false positive than the \textbf{real detector}.

\section{Conclusions}
\label{sec:conclusions}
In this paper we proposed a novel GAN architecture, based on LSGAN, to
transform visible spectrum images in thermal spectrum ones. We also proposed
a novel training procedure that mixes real and synthesized images to adapt the
YOLOv3 detector for detection in the thermal domain. Extensive experimental
validation shows that our method outperforms state-of-the-art
single-modality detectors for pedestrian detection on the KAIST dataset.

Our experiments show that that even using only 50\% of available real thermal
images it is possible to obtain results that are comparable with
state-of-the-art methods trained using 100\% real thermal images. This suggests
that images generated with our proposed GAN are beneficial and may help to adapt
visible spectrum detectors to operate in thermal spectrum in domains suffering
from a lack of training data.

\paragraph*{Acknowledgement}
This research was partially funded by Leonardo.\vspace{-8pt}



\bibliographystyle{IEEEtran}
\bibliography{thermal-gan}

\end{document}